\documentclass[pdflatex,sn-mathphys-num]{sn-jnl}

\usepackage[table,xcdraw]{xcolor}
\usepackage{graphicx}%
\usepackage{multirow}%
\usepackage{amsmath,amssymb,amsfonts}%
\usepackage{amsthm}%
\usepackage{mathrsfs}%
\usepackage[title]{appendix}%
\usepackage{xcolor}%
\usepackage{textcomp}%
\usepackage{manyfoot}%
\usepackage{booktabs}%
\usepackage{algorithm}%
\usepackage{algorithmicx}%
\usepackage{algpseudocode}%
\usepackage{listings}%
\usepackage{booktabs}
\usepackage{multirow}
\usepackage{graphicx}
\usepackage{array}
\usepackage{caption}
\usepackage{lmodern}  
\usepackage{rotating}
\usepackage{lscape}

\theoremstyle{thmstyleone}%
\usepackage[table,xcdraw]{xcolor}
\usepackage{longtable}
\usepackage{colortbl}
\usepackage{graphicx}
\theoremstyle{thmstyletwo}%

\theoremstyle{thmstylethree}%

\usepackage{lscape}
\begin{document}

\title[Article Title]{SaudiCulture: A Benchmark for Evaluating Large Language Models' Cultural Competence within Saudi Arabia.}


\author*[1]{\fnm{Lama} \sur{Ayash}}\email{444800100@kku.edu.sa}

\author[2]{\fnm{Hassan }\sur{Alhuzali}}\email{hrhuzali@uqu.edu.sa}

\author[1, 3]{\fnm{Ashwag} \sur{Alasmari}}\email{aasmry@kku.edu.sa}
\author[4]{\fnm{Sultan} \sur{Aloufi}}\email{sdoufi@taibahu.edu.sa}

\affil*[1]{\orgdiv{Department of Computer Science}, \orgname{King Khalid University}, \orgaddress{\city{Abha}, \country{Saudi Arabia}}}

\affil[2]{\orgdiv{Department of Computer Science \& Artificial Intelligence}, \orgname{Umm Al-Qura University}, \orgaddress{\city{Makkah}, \country{Saudi Arabia}}}

\affil[3]{\orgdiv{Center for Artificial Intelligence (CAI)}, \orgname{King Khalid University}, \orgaddress{\city{Abha}, \country{Saudi Arabia}}}

\affil[4]{\orgdiv{Department of Arabic Language}, \orgname{Taibah University}, \orgaddress{\city{Medina}, \country{Saudi Arabia}}}

\abstract{Large Language Models (LLMs) have demonstrated remarkable capabilities in natural language processing; however, they often struggle to accurately capture and reflect cultural nuances. This research addresses this challenge by focusing on Saudi Arabia, a country characterized by diverse dialects and rich cultural traditions. We introduce SaudiCulture, a novel benchmark designed to evaluate the cultural competence of LLMs within the distinct geographical and cultural contexts of Saudi Arabia. SaudiCulture is a comprehensive dataset of questions covering five major geographical regions—West, East, South, North, and Center—along with general questions applicable across all regions. The dataset encompasses a broad spectrum of cultural domains, including food, clothing, entertainment, celebrations, and crafts. To ensure a rigorous evaluation, SaudiCulture includes questions of varying complexity, such as open-ended, single-choice, and multiple-choice formats, with some requiring multiple correct answers. Additionally, the dataset distinguishes between common cultural knowledge and specialized regional aspects. We conduct extensive evaluations on five LLMs—GPT-4, Llama 3.3, FANAR, Jais, and AceGPT—analyzing their performance across different question types and cultural contexts. Our findings reveal that all models experience significant performance declines when faced with highly specialized or region-specific questions, particularly those requiring multiple correct responses. Furthermore, we observe that while some regions are better understood by LLMs, others remain largely misrepresented. For instance, GPT-4 achieves the highest accuracy in the western region (66\%), whereas Jais records the lowest accuracy in the northern region (16\%). Additionally, certain cultural categories are more easily identifiable than others, further highlighting inconsistencies in LLMs’ cultural understanding. These results emphasize the importance of incorporating region-specific knowledge into LLMs training to enhance their cultural competence. We hope that SaudiCulture serves as a foundation for future efforts aimed at improving the ability of LLMs to engage with and accurately represent diverse cultural contexts.}




\maketitle

\section{Introduction}\label{sec:introduction}

Large Language Models (LLMs) have demonstrated remarkable progress in natural language processing (NLP), achieving state-of-the-art performance on a wide range of tasks, including text generation, translation, and question answering. However, their success is often contingent on the availability of massive datasets, which are predominantly comprised of data in high-resource languages like English. The majority of LLMs have been predominantly trained on data from high-resource languages like English, leading to an over-reliance on West perspectives and a limited understanding of diverse cultural contexts. 

Culture is a multifaceted concept that encompasses the way of life, including our thoughts and actions \cite{liu2024culturally}. It encompasses both tangible elements such as food, art and clothing, and intangible aspects such as ideas, values, attitudes, and norms. Culture is also significantly shaped by geographical location, ethnicity, and history. In Saudi Arabia, for example, cultural norms vary significantly between different regions \cite{long2005culture}.In the central region of Saudi Arabia, traditional celebrations often feature Al-Ardah, a historic Najdi sword dance performed to rhythmic drumbeats and poetic chants. Participants wear the iconic daglah, a long embroidered coat, along with a ghutra and agal. In contrast, coastal regions such as East or West may have a stronger influence on maritime traditions, with a focus on fishing and a distinct coastal cuisine. On the other hand, Southern region boasts a distinct cultural heritage marked by ancient traditions, unique architectural styles, and vibrant folk arts. Similarly, The Northern region of Saudi Arabia is renowned for its rich tradition of olive cultivation, which is home to some of the oldest and largest olive farms in the Kingdom. The Kingdom of Saudi Arabia boasts a unique blend of Bedouin heritage, urban modernity, and religious and social customs. This cultural diversity presents both a unique opportunity and a significant challenge for LLMs. 

\begin{figure}[ht]
    \centering
    \includegraphics[width=1.0\textwidth]{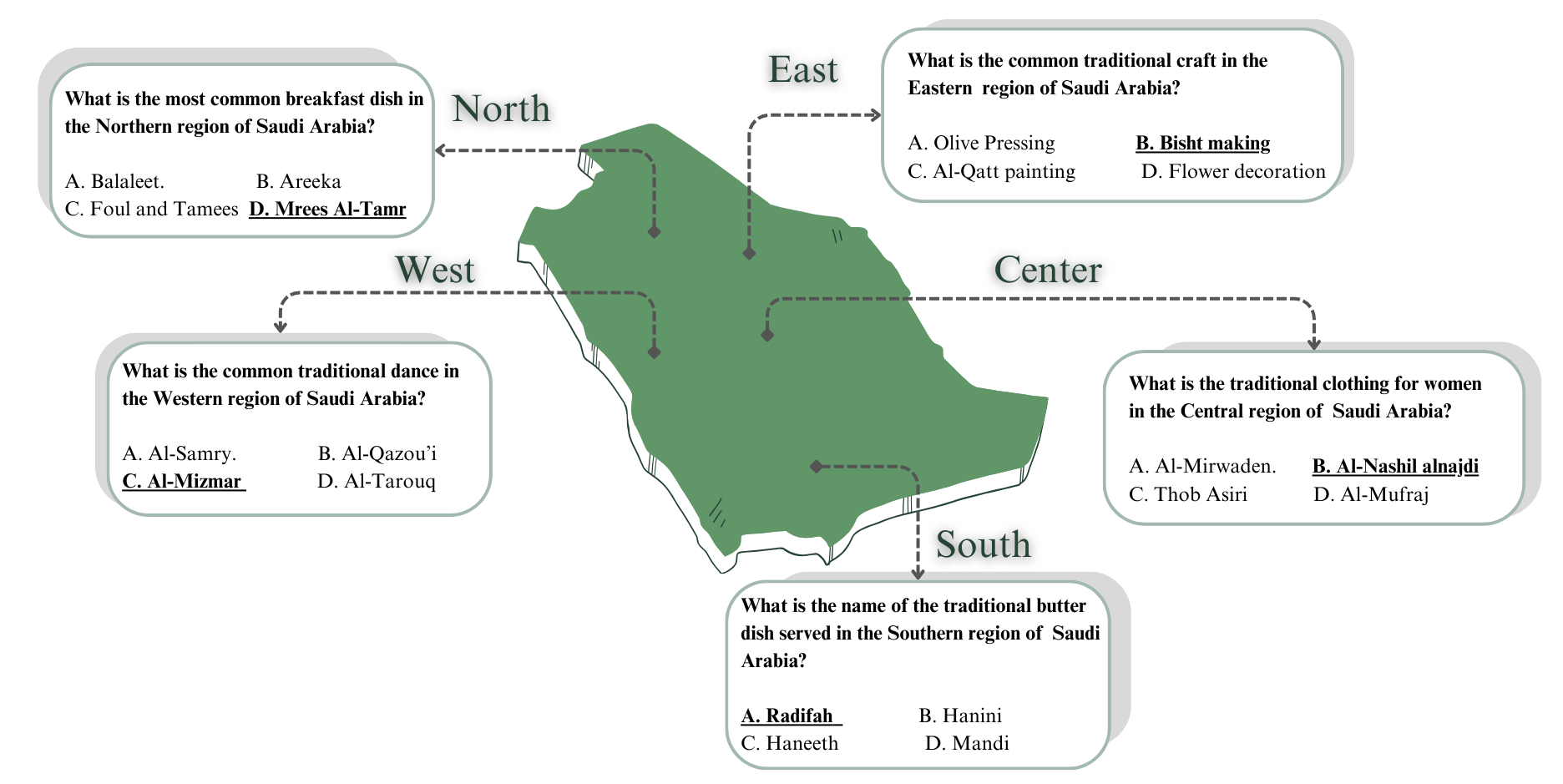}
    \caption{Regional cultural diversity in Saudi Arabia, showcasing traditional food, crafts, and clothing across different areas. This highlights the rich cultural landscape that LLMs must understand to provide contextually relevant responses}
    \label{fig:map}
\end{figure}

To evaluate the performance of LLMs within this context of Saudi Arabian culture, we have developed a novel benchmark dataset specifically designed to assess their understanding of Saudi culture, as some examples are shown in Figure \ref{fig:map} from the SaudiCulture dataset. This dataset also encompasses a wide range of questions covering five major regions of the Kingdom: South, North, West, East, and Central, along with common questions across all regions. In this paper, our goal is to achieve the following objectives:
\begin{itemize}

\item Establish a benchmark SaudiCulture that reflects the cultural diversity of Saudi Arabia, covering five major regions and multiple cultural domains. 
\item Evaluate the performance of state-of-the-art LLMs on tasks involving culturally nuanced and linguistically diverse data through extensive analysis.

\item Investigate the impact of different question formats (open-ended, single-choice, and multiple-choice) on the accuracy of LLMs' cultural competence.

\item Identify strengths and limitations in the models' understanding of Saudi-specific cultural nuances and highlight areas for improvement.

\end{itemize}
Building on these objectives, our study makes three primary contributions. First, it introduces SaudiCulture, a culturally rich benchmark dataset covering five major regions of Saudi Arabia, with common questions representing all five regions along with general questions applicable across all regions. It also includes various aspects such as food, clothing, and celebrations, among others. Second, it provides an empirical assessment of five cutting-edge LLMs, examining their responses across different question types: open-ended, single-choice, and multiple-choice to offer a nuanced view of cultural competence. Third, it presents a detailed performance analysis of these models when faced with region-specific and multi-answer queries, illuminating existing knowledge gaps and suggesting potential refinements in cultural understanding\footnote{Upon acceptance, we will make SaudiCulture available to the research community.}.

The paper is structured as follows: Section \ref{relatedwork} reviews related work, while Section \ref{data} details the data collection process. Section \ref{Experiment} describes the experimental setup and results, followed by Section \ref{analysis}, which presents the analysis. Section \ref{discussion} provides the discussion, including limitations and ethical considerations. Finally, Section \ref{conc} concludes with a summary of key insights and directions for future research.

\section{Related Work}\label{relatedwork}
LLMs while demonstrating remarkable proficiency in various NLP tasks, often fall short when it comes to a nuanced understanding and reflection of diverse cultures.  Their strong capabilities in areas such as text generation, translation, and question answering are undermined by a fundamental challenge: effectively capturing the intricate interplay of language, social norms, values, and traditions that constitute culture. This deficiency raises concerns about the potential of LLMs to perpetuate biases, misinterpret cultural cues, and ultimately fail to engage with users in a culturally sensitive and appropriate manner. 
\subsection{Evaluation of LLMs for Cultural Understanding}
The impact of this cultural gap in LLMs is significant.  As LLMs become increasingly integrated into various aspects of human life, their influence on individuals and societies grows.  Research has shown that LLMs can shape opinions, provide persuasive advice, and even influence perceptions of personality \cite{griffin2023susceptibility, buchanan2021truth, rao2023can, vodrahalli2022humans}. However, the limited diversity of thought exhibited by these models \cite{park2024diminished}, particularly in the realm of cultural understanding, presents the risk of reinforcing existing stereotypes and promoting a homogenized, often Western-centric, worldview.  This demonstrates the urgent need for a rigorous evaluation and a targeted improvement in the cultural competency of LLMs.

Existing research has begun to explore the complexities of LLM cultural understanding. Some studies have focused on identifying inherent limitations, such as the tendency to rely on simplified heuristics rather than nuanced reasoning \cite{shapira2023clever}, inconsistencies in moral and legal judgments in different contexts \cite{almeida2023exploring}, and a general lack of sensitivity to social cues \cite{gandhi2024understanding}. Researchers have also used psychological frameworks to analyze the personalities and values of LLMs, revealing potential biases and raising ethical concerns \cite{li2022gpt, dhingra2023mind, binz2023using}.  These investigations highlight the multifaceted nature of cultural understanding and the difficulty of replicating human-like social intelligence in artificial systems.

Efforts are also underway to enhance LLM cultural sensitivity. Approaches like EmotionPrompt \cite{li2023emotionprompt} and the work by Safdari et al. \cite{safdari2023personality} explore the potential of prompting and fine-tuning to modulate LLMs responses and align them more closely with human emotional and social expectations. However, these initiatives primarily focus on individual emotions and personality traits, leaving the broader domain of cultural competency largely unexplored.

Direct analysis of cultural dimensions in LLMs has been relatively limited.  Although studies utilizing Hofstede's cultural dimensions \cite{arora2022probing, wang2023cdeval, kharchenko2024well} have provided valuable insights into cross-cultural biases, they often focus on Western LLMs and rely heavily on closed-ended question formats.  Furthermore, the scarcity of research on multilingual LLMs and their cultural understanding across different languages and dialects represents a significant gap.

More recent work has begun to address these limitations. Researchers are exploring alternative frameworks for cultural evaluation, such as survey-based approaches \cite{ramezani2023knowledge, alkhamissi2024investigating, warmsley2024assessing}, and developing new benchmarks like CulturalBench \cite{chiu2024culturalbench} to assess cultural knowledge and reasoning in more nuanced ways.  The extension of cultural benchmarks to vision-language models \cite{nayak2024benchmarking, zhang2024cultiverse} further demonstrates the growing attention of the importance of cultural understanding in Artificial Intelligence (AI) systems.

Despite these advancements, crucial gaps remain.  As pointed out by Adilazuarda et al. \cite{adilazuarda2024towards}, the field still needs more research that leverages multi-lingual datasets and grounds its investigations in robust social science frameworks.  Furthermore, the emphasis on cultural knowledge often overshadows the crucial aspect of cultural adaptability – the ability to apply cultural understanding in contextually appropriate ways.  Existing efforts to improve adaptability, such as synthetic persona generation \cite{alkhamissi2024investigating, kwok2024evaluating, durmus2023towards}, represent a promising direction but require further refinement.  The counterintuitive finding that cross-lingual probing does not always enhance performance \cite{durmus2023towards, shen2024understanding}, which highlights the complexities of cultural understanding in LLMs and warrants further investigation.

\subsection{Existing Cultural Benchmarks}

Evaluating the cultural competence of LLMs requires robust and nuanced benchmarks. Even though the field is still relatively nascent, several approaches have emerged, each with its strengths and limitations. These benchmarks can be broadly categorized by their focus and methodology, including those based on established cultural frameworks, survey-based approaches, and those employing novel question formats. 

Early efforts, like Anacleto et al. \cite{anacleto2006can}, focused on collecting commonsense knowledge about specific cultural practices (e.g., eating habits) through crowd-sourcing platforms. GeoMLAMA \cite{yin2022geomlama} broadened the scope by incorporating geo-diverse commonsense concepts across five countries, using native languages. Nguyen et al. \cite{nguyen2023extracting} explored leveraging large web corpora like Common Crawl to extract cultural commonsense knowledge related to geography, religion, and occupations. Datasets like CREHate \cite{lee2024exploring} addressed cross-cultural differences in hate speech, while CultureAtlas \cite{fung2024massively} compiled textual data from Wikipedia to represent cultural norms across numerous countries. However, these early benchmarks primarily focused on English data and often relied on formal data sources, potentially missing the everyday nuances of cultural expression.  

More recent work has shifted towards evaluating LLMs on non-English languages and cultures. Studies like CLIcK \cite{kim2024click} and HAE-RAE Bench \cite{son2023hae} assessed LLMs knowledge in Korean, while COPAL-ID \cite{wibowo2023copal}, ID-CSQA \cite{putri2024can}, and IndoCulture \cite{10.1162/tacl_a_00726} incorporated culturally nuanced questions in Indonesian. However, a direct comparison of LLMs cultural adaptiveness across diverse languages using a standardized question set has been lacking. 

Other research has explored the capture of everyday cultural nuances through data from social media platforms. StereoKG \cite{deshpande2022} extracted cultural stereotypes from X (formerly Twitter) and Reddit, while CAMEL \cite{naous2024} focused on Arabic content from X, and CultureBank \cite{shi2024} collected diverse perspectives on cultural descriptors from TikTok and Reddit. Although these approaches offer insights into culturally relevant online discourse, they are limited by their reliance on a single platform and the potential for noise and bias in social media data. Furthermore, they may not fully capture the full spectrum of everyday cultural behaviors. 

Recognizing the importance of culturally sensitive LLMs for the Arabic-speaking world, several benchmarks have been developed. Palm \cite{alwajih2025palmculturallyinclusivelinguistically} introduces a culturally inclusive and linguistically diverse dataset covering Arab countries, emphasizing both Modern Standard Arabic (MSA) and dialectal Arabic. ArabCulture \cite{sadallah2025commonsense} addresses the limitations of machine-translated datasets by providing a commonsense reasoning dataset in MSA, constructed by native speakers across different Arab countries. AraDiCE \cite{mousi2024aradice} focuses on dialectal and cultural capabilities in LLMs, offering benchmarks for dialect comprehension and generation, particularly for low-resource Arabic dialects, and introduces a fine-grained cultural awareness benchmark. CIDAR \cite{alyafeai2024cidar} further emphasizes cultural relevance by providing an Arabic instruction-tuning dataset culturally aligned by native Arabic speakers, addressing the biases inherent in English-dominated datasets. These efforts collectively highlight the growing recognition of the need for culturally adapted resources for Arabic LLMs. Furthermore, a critical gap exists: no comprehensive benchmark currently addresses the rich and diverse cultural landscape of Saudi Arabia. Given the distinct regional variations, traditions, and social norms that define Saudi culture, a dedicated and culturally sensitive evaluation framework is essential. To bridge this gap, we introduce SaudiCulture, a novel benchmark designed to assess the cultural competence of LLMs in the context of Saudi Arabia. Figure \ref{fig:map} presents a selection of questions from our data, illustrating the regional cultural diversity in Saudi Arabia.
\section{Data Collection}\label{data}
\begin{figure}[ht]
    \centering
    \includegraphics[width=1.0\textwidth]{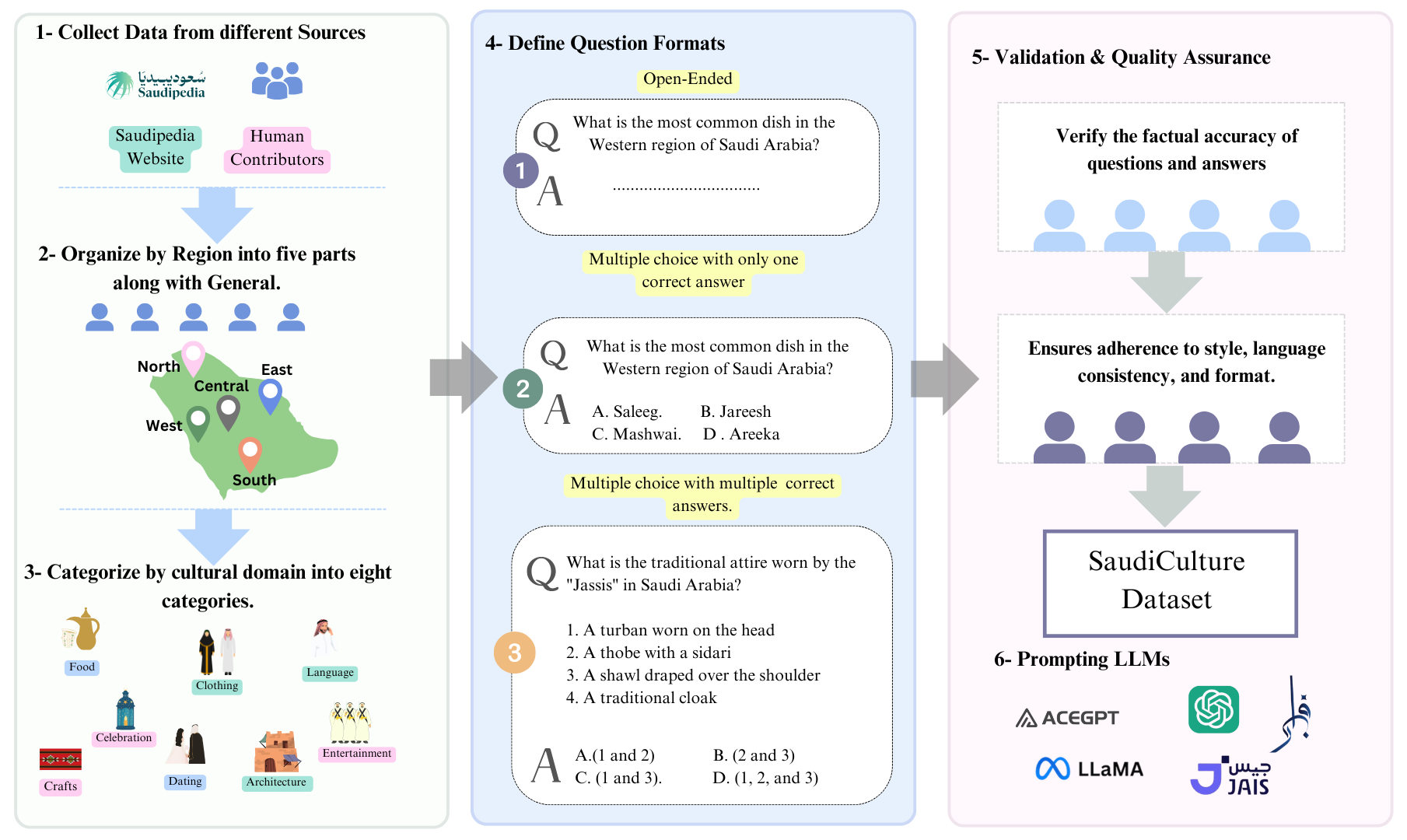}
    \caption{Overview of the data collection methodology for SaudiCulture. The process starts with gathering information from diverse sources and organizing the questions by regions and cultural categories. Questions are then assigned one of three formats. The result is the SaudiCulture dataset, comprising 441 culturally rich questions.}
    \label{fig:methodology}
\end{figure}

We introduce SaudiCulture, a comprehensive dataset designed to evaluate the ability of LLMs to understand and respond to culturally specific questions related to Saudi Arabia. The creation of this dataset involved a rigorous process of curation, validation, and organization to ensure both cultural authenticity and linguistic relevance. SaudiCulture includes $441$ questions along with their corresponding answers, carefully sourced from a diverse set of references, including a reputable online platform focused on Saudi culture, called Saudipedia\footnote{\url{https://saudipedia.com/}}, and direct contributions from human experts familiar with the cultural context. The primary objective of SaudiCulture is to serve as a valuable benchmark for assessing the cultural competency of LLMs, measuring their ability to accurately interpret and respond to culturally grounded queries. SaudiCulture spans multiple cultural domains, capturing regional variations, categorical diversity, and contextual richness to ensure comprehensive coverage. In addition to covering diverse cultural aspects, SaudiCulture also incorporates various response formats, enabling a more granular analysis of LLM performance across different types of cultural knowledge from factual historical events to social norms and linguistic expressions. This structured design provides a robust and challenging framework for evaluating how well LLMs grasp the subtle and often implicit elements of Saudi culture. A visual summary of the data collection and curation process is presented in Figure~\ref{fig:methodology}.

\subsection{Regional Diversity in Data Collection} 

For a comprehensive representation of Saudi Arabian cultural contexts, the SaudiCulture dataset incorporates questions from five geographically distinct regions: north, south, west, east, and central. Each region contributes region-specific questions that reflect its unique cultural elements, such as local traditions, historical events, dialectal expressions, and social practices. This regional segmentation allows for a detailed evaluation of LLMs' ability to recognize, interpret, and appropriately respond to culturally nuanced content rooted in specific geographical contexts.

In addition to these region-specific questions, SaudiCulture  includes a separate group labeled "general." The general group contains questions that are culturally relevant across all regions, representing shared national heritage, customs, and knowledge that are common to Saudi Arabia as a country. These general questions serve two important purposes: (1) They provide a baseline for evaluating general cultural competency, independent of regional variations of Saudi Arabia. (2) They help assess whether LLMs can differentiate between localized cultural elements and those that are universally understood across the country of Saudi Arabia. By combining regional diversity with a nationally shared cultural core, SaudiCulture offers a balanced and holistic evaluation framework, ensuring that models are tested for both local cultural sensitivity and broad cultural awareness.



\subsection{Diversity by Cultural Categories} 

The SaudiCulture dataset covers a wide range of cultural categories, including food, clothing, language, entertainment, celebrations, dating, crafts, and architecture cultural aspects. These categories, adapted from Koto et al. \cite{koto2024indoculture} to reflect Saudi general and regional cultural practices, are detailed in Table~\ref{tab: tab 1}. Each category is represented by a dedicated set of questions designed to probe specific cultural practices, traditions, artifacts, and expressions associated with that domain. For example, questions within the food category explore regional cuisines, signature dishes, and culinary customs, while the celebration category focuses on traditional instruments, folk songs, and the cultural significance of particular musical forms. Similarly, the clothing category covers traditional attire and its variations across different regions and occasions. This categorical diversity ensures that the dataset provides a well-rounded and multidimensional view of Saudi culture, allowing for a thorough evaluation of LLMs' ability to engage with cultural knowledge across diverse domains.

\begin{table}[]
\centering
\caption{Overview of cultural categories and the sample counts for each category across the five geographical regions along with the general. }
\label{tab: tab 1}
\begin{tabular}{l|cccccc|c}
\toprule
Categories/Reginal & Center & East & General & North & South & West & Total\\ \midrule
Celebration        &      4  &   6   &     6    &   2    &    4   &   8 & 30  \\
Clothing           &     4   &   7   &     6    &   4    &    6   & 5 & 32    \\
Craft \& work             &   11     &   19   &     3    &  2     & 16      &   9  & 60  \\
Entertainment      &     21   &   10   &    21   &   14    &   15    & 14  & 95   \\
Food               &    24    &   19   &     9    &   23    &   26    &   24 & 125  \\
Architecture    &   6     &   2   &     5    &   1   &  9     &  0  & 23  \\
Language \& Communication   &    2    &   0   &   32     &    2   & 2      &   4 & 42  \\
Dating   &    4    &    4   &   4      &    6  &   6    &   10  & 34 \\
  \midrule
Total            &   76     &    67  &    86     &    54   &  84     & 74 & 441\\
\bottomrule
\end{tabular}
\end{table}



\subsection{Differentiation by Levels of Complexity} 

SaudiCulture  incorporates questions across three distinct answer formats, as summarized in Table~\ref{tab:my-table}. The first format consists of open-ended questions, where the model is given full freedom to generate responses without predefined constraints. This format tests the model’s ability to construct coherent, contextually relevant, and culturally appropriate responses using its internal knowledge. The second format uses single-answer multiple-choice questions, requiring the model to select the most accurate response from a predefined set of options. This format evaluates the model’s precision in factual recall and cultural disambiguation. The third format features multiple-answer multiple-choice questions, where more than one option is correct. This format challenges the model’s ability to recognize all valid answers, requiring a deeper understanding of cultural nuances and interrelated knowledge. 

Furthermore, our data includes two distinct question types: common and specialized cultural questions. Common questions focus on widely recognized cultural elements, such as food, dance, craft, and other cultural elements. Specialized cultural questions, on the other hand, pertain to specific regional practices, such as ``What is the traditional clothing of Tehama?''. Tehama is geographically located in the southern region of Saudi Arabia, making the question is more specific to the region. By incorporating these varied answer types and question categories, the dataset ensures a comprehensive evaluation framework, enabling analysis of both generative and discriminative cultural competence across different levels of cognitive and cultural complexity.

\begin{table}[ht]
\centering
\caption{Overview of type of answers formats and the sample counts for each type across the five regions and general question.}
\label{tab:my-table}
\begin{tabular}{l|cccccc|c}
\toprule
Resonces/Reginal & Center & East & General & North & South & West & Total\\ \midrule
Open-ended     &    31    &  29    &    50     &  24 &38    &   34  & 206     \\
Single answer - MCQ       &    31    &   29   &   19      & 24 & 38    &   34   & 175     \\
Multiple answers - MCQ        &    14    &  9    &     15 & 6  &    10   &   6   & 60      \\ \midrule
Total          &    76    &   67   &     84 & 54    &    86   &    74   & 441  \\ \bottomrule    
\end{tabular}
\end{table}

\subsection{Validation of Data Collection}

To ensure the accuracy, reliability, and cultural authenticity of SaudiCulture, we employed a rigorous multi-stage validation process that combined source verification (i.e., relying on a rebuttable online platform), peer review, regional expert feedback, and expert linguistic validation. This triangulation step ensured that the factual content used to construct questions and answers was consistent and verifiable. To further enhance cultural authenticity, we engaged native reviewers from each of the five geographical regions represented in the dataset. These cultural informants reviewed the questions and answers to verify cultural relevance and contextual accuracy. Their feedback helped to correct discrepancies and refine content to ensure it accurately reflected the lived cultural experiences and traditions of each region. An expert holding a Ph.D. conducted an in-depth linguistic review of the content. This expert ensured that the text adhered to correct grammar, syntax, and cultural expression, maintaining both linguistic accuracy and cultural fidelity. Together, these multi-level validation measures — including source triangulation, peer review, regional cultural input, and expert linguistic assessment — ensured that SaudiCulture is not only factually accurate but also culturally authentic and linguistically sound, making it a robust resource for evaluating cultural competence in LLMs.




\section{Experiments}\label{Experiment}
\begin{figure}[ht]
    \centering
    \includegraphics[width=0.9\textwidth]{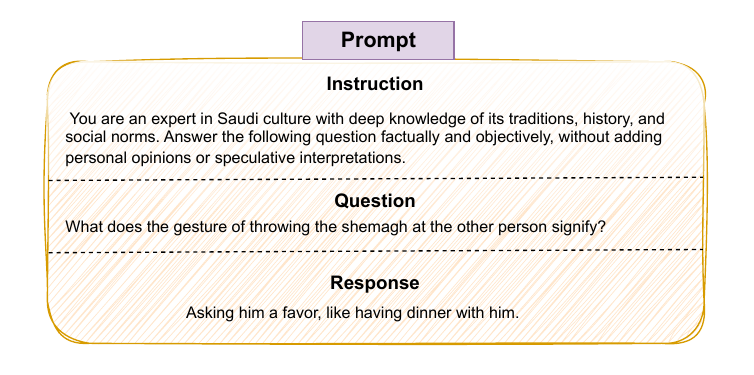}
    \caption{Showing an illustration of the prompt used in this work.}
    \label{fig:prompt}
\end{figure}

This study assesses the effectiveness of LLMs in understanding and generating responses concerning SaudiCulture. Five LLMs were chosen based on their training data and linguistic capabilities. Among these models, GPT-4~\cite{openai2024gpt4technicalreport} and Llama-3.3~\cite{grattafiori2024llama3herdmodels} stand out for being trained on diverse linguistic datasets, including Arabic, enabling them to excel in cross-lingual assessments. Moreover, the study includes three models mainly trained on Arabic data: Fanar~\cite{fanarteam2025fanararabiccentricmultimodalgenerative}, Jais~\cite{sengupta2023jaisjaischatarabiccentricfoundation}, and AceGPT-32B~\cite{huang-etal-2024-acegpt}. By incorporating both multilingual and Arabic-centric models, we can uncover their respective strengths and weaknesses in processing culturally and linguistically intricate Arabic content. We also experiment with three types of questioning the LLMs, an illustration is shown in Figure~\ref{fig:prompt}. The initial type allows the model to freely generate responses, providing insight into its baseline performance. The subsequent type evaluates its precision in selecting only one correct response from multiple choices. The final type assesses its capacity to identify and select multiple correct responses when relevant. These three approaches facilitate a thorough comparison of the model performance in various interaction scenarios.

To evaluate the results of generated responses when answering the questions, accuracy serves as the primary metric, validating whether the response generated by the model includes at least one correct answer. When dealing with multiple-choice options, accuracy is verified by assessing the model's selection of the correct option from the available choices. In cases where multiple correct answers are applicable, the response is considered accurate only if all relevant answers are correctly identified. This evaluation ensures a comprehensive assessment of the LLMs' ability to understand cultural aspects at different levels and types.

\subsection{Experimental Results}
The results shown in Table~\ref{tab:results} highlight GPT-4's exceptional performance, surpassing even the Arabic models in this cultural comprehension task. Moreover, a key observation from the results is the challenging nature of open-ended questions compared to closed-ended ones. This finding underscores the complexity inherent in answering open-ended queries, which can have a multitude of possible responses, posing a greater challenge for LLMs. Such questions, by their nature, lack specificity and structure, making it more difficult for models to provide accurate responses, especially considering the variations in responses that could arise based on regional and contextual differences. LLMs also exhibit a higher level of cultural understanding in the western region compared to the northern region. This disparity can be attributed to the extensive cultural interactions that characterize the western region, particularly due to the significant presence of pilgrims and visitors in Makkah and Medina. The continuous exposure to diverse cultural perspectives and backgrounds in these revered cities likely enhances the cultural acumen of the local population, reflecting the region's enriched cultural environment and its impact on the models' comprehension abilities. 

\begin{table}[h]
\centering
\caption{Accuracy percentages of five LLMs on cultural knowledge tasks across regions in Saudi Arabia. The table distinguishes between common and specialized queries and details performance for each question format.}
\label{tab:results}
\begin{tabular}{lllccccc}
\toprule
\textbf{Region} & \textbf{Query Type} & \textbf{Question Format} & \textbf{GPT-4} & \textbf{Llama} & \textbf{FANAR} & \textbf{Jais} & \textbf{AceGPT} \\
\midrule
\multirow{5}{*}{West} 
   & \multirow{2}{*}{Common}      & Open-Ended       & 54\%  & 23\%  & 15\%  & 15\%  & 23\% \\
   &                             & Single-Answer    & 77\%  & 38\%  & 54\%  & 31\%  & 38\% \\
   & \multirow{3}{*}{Specialized}  & Open-Ended       & 52\%  & 28\%  & 5\%   & 5\%   & 0\%  \\
   &                             & Single-Answer    & 81\%  & 62\%  & 43\%  & 52\%  & 47\% \\
   &                             & Multi-Answer     & 77\%  & 38\%  & 54\%  & 31\%  & 38\% \\ \midrule 
   & Overall     & & 66\%  & 39\%  & 25\%  & 24\%  & 25\% \\ 

\midrule
\multirow{5}{*}{South} 
   & \multirow{2}{*}{Common}      & Open-Ended       & 54\%  & 15\%  & 7\%   & 7\%   & 7\%  \\
   &                             & Single-Answer    & 77\%  & 23\%  & 38\%  & 7\%   & 30\% \\
   & \multirow{3}{*}{Specialized}  & Open-Ended       & 40\%  & 56\%  & 12\%  & 24\%  & 28\% \\
   &                             & Single-Answer    & 77\%  & 23\%  & 38\%  & 7\%   & 30\% \\
   &                             & Multi-Answer     & 70\%  & 40\%  & 10\%  & 30\%  & 40\% \\
\midrule 
& Overall     & & 56\%  & 44\%  & 26\%  & 22\%  & 38\% \\ \midrule 

\multirow{5}{*}{East} 
   & \multirow{2}{*}{Common}      & Open-Ended       & 38\%     & 15\%    & 0\%    & 0\%    & 15\% \\
   &                             & Single-Answer    & 46\%  & 38\% & 30\%& 15\%   & 30\% \\
   & \multirow{3}{*}{Specialized}  & Open-Ended       & 50\%     & 12\%    & 0\%    & 6\%    & 12\% \\
   &                             & Single-Answer    & 75\%     & 68\%    & 18\%   & 31\%   & 50\% \\
   &                             & Multi-Answer     & 66\%     & 66\%    & 55\%   & 44\%   & 22\% \\
\midrule
& Overall     & & 52\%  & 41\%  & 17\%  & 17\%  & 27\% \\ \midrule
\multirow{5}{*}{Center} 
   & \multirow{2}{*}{Common}      & Open-Ended       & 46\%  & 23\%  & 23\%  & 15\%  & 15\% \\
   &                             & Single-Answer    & 77\%  & 62\%  & 46\%  & 46\%  & 54\% \\
   & \multirow{3}{*}{Specialized}  & Open-Ended       & 41\%  & 12\%  & 0\%   & 0\%   & 12\% \\
   &                             & Single-Answer    & 53\%  & 41\%  & 35\%  & 41\%  & 29\% \\
   &                             & Multi-Answer     & 42\%  & 32\%  & 14\%  & 35\%  & 35\% \\
\midrule
& Overall     & & 51\%  & 34\%  & 25\%  & 27\%  & 27\% \\ \midrule
\multirow{5}{*}{North} 
   & \multirow{2}{*}{Common}      & Open-Ended       & 15\%  & 23\%  & 8\%   & 8\%   & 8\%  \\
   &                             & Single-Answer    & 38\%  & 31\%  & 38\%  & 31\%  & 31\% \\
   & \multirow{3}{*}{Specialized}  & Open-Ended       & 63\%  & 36\%  & 27\%  & 18\%  & 27\% \\
   &                             & Single-Answer    & 36\%  & 63\%  & 45\%  & 0\%   & 55\% \\
   &                             & Multi-Answer     & 20\%  & 0\%   & 40\%  & 20\%  & 20\% \\ \midrule
& Overall     & & 36\%  & 34\%  & 30\%  & 16\%  & 28\% \\ 
\bottomrule
\end{tabular}%
\end{table}

\section{Analysis}\label{analysis}
This section provides a comprehensive overview of the evaluation of LLMs on the SaudiCulture. We examine how these models process and respond to culturally nuanced queries, focusing on both general and region-specific questions across various regions of Saudi Arabia. By analyzing performance variations in response to different question formats and the influence of training data representation, we aim to uncover the underlying factors that drive accuracy discrepancies.

\subsection{General and Regional Cultural Understanding} 
\begin{figure}[ht]
    \centering
    \includegraphics[width=\textwidth]{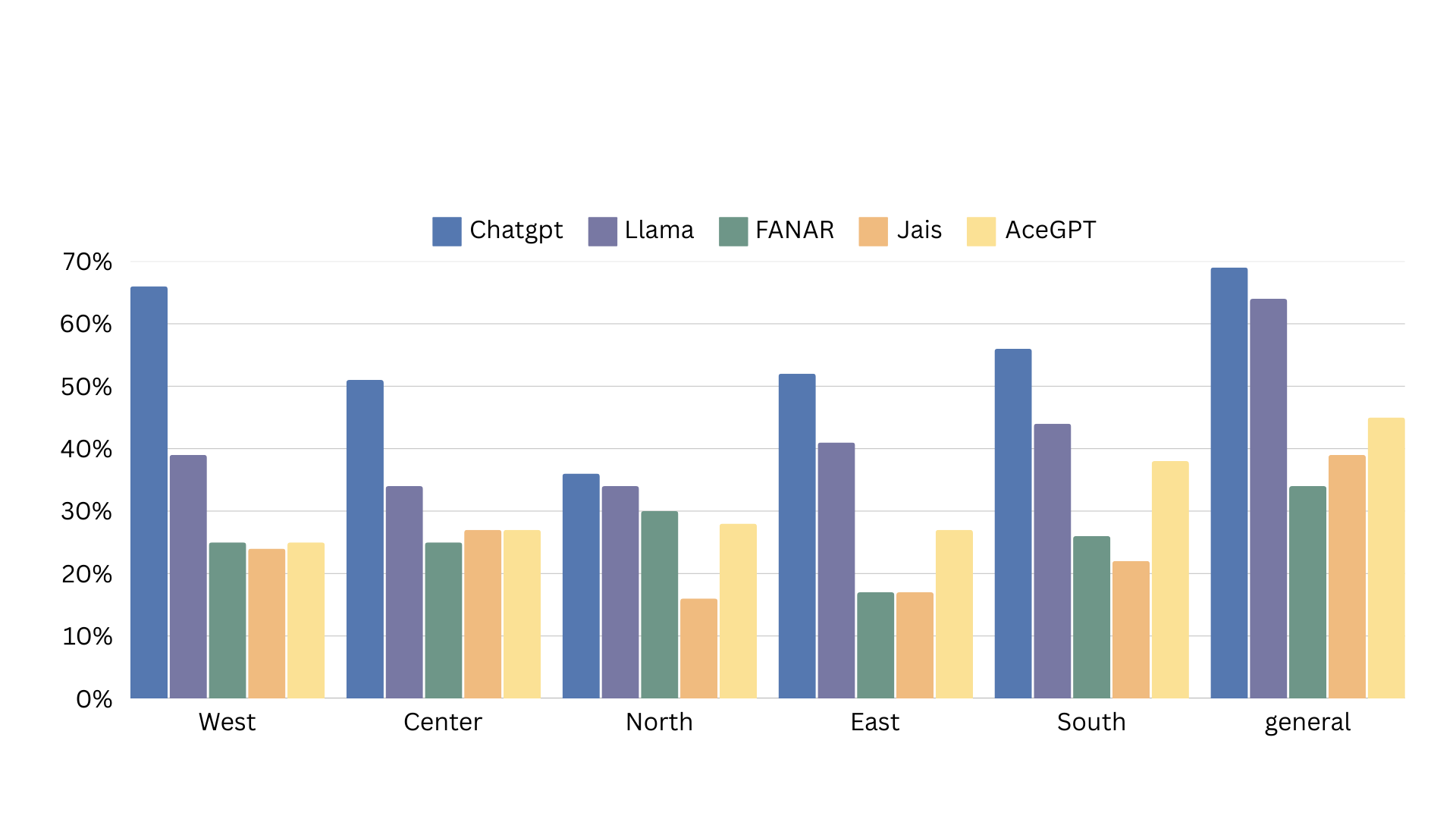}
    \caption{Model accuracy across the five regions and general cultural questions. Each region is represented with a distinct color, and the bars display the performance of the five models: GPT-4, Llama, FANAR, Jais, and AceGPT. This visualization provides an overview of how well each model performs in handling culturally specific and general queries.}
    \label{fig:accuracy}
\end{figure}

When evaluating performance on general questions, those that are common across all regions, all models demonstrate notably better performance compared to their region-specific results, as seen in Figure~\ref{fig:accuracy}. This disparity demonstrates the inherent complexity associated with capturing fine-grained region-specific cultural knowledge, highlighting the importance of incorporating localized and culturally grounded approaches to enhance model competency in culturally diverse contexts. The chosen models consistently perform better in capturing general cultural contexts of Saudi Arabia compared to region-specific nuances.

For example, GPT-4 achieved the highest accuracy on general cultural data at 69\%, but its performance dropped considerably for regional data, with 36\% in the north and 52\% in the east, respectively. Similarly, Llama exhibited  strong accuracy at 64\% for general, but its regional scores ranged from 34\% to 44\%. Arabic models like FANAR, Jais, and AceGPT followed the same pattern, with their overall accuracies (34\%, 39\%, and 45\%, respectively) exceeding their performance on region-specific questions.

An analysis of the five models' performance across the five Saudi regions reveals several significant trends. As shown in Figure~\ref{fig:accuracy}, GPT-4 demonstrates superior accuracy across multiple regions, achieving an overall accuracy of 66\% in the western region and 56\% in the southern region. In contrast, Llama exhibits moderate yet stable performance, with regional accuracies consistently ranging between 30\% and 40\%. FANAR and Jais, however, display more substantial performance declines when confronted with region-specific questions, with accuracy occasionally dropping as low as 17\% for certain questions from the eastern region. AceGPT’s performance generally falls within a mid-range, achieving accuracies between 25\% and 45\% across the five regions.

\subsection{Option-Provided Versus Unconstrained Queries} 

\begin{figure}[ht]
    \centering
    \includegraphics[width=1.0\textwidth]{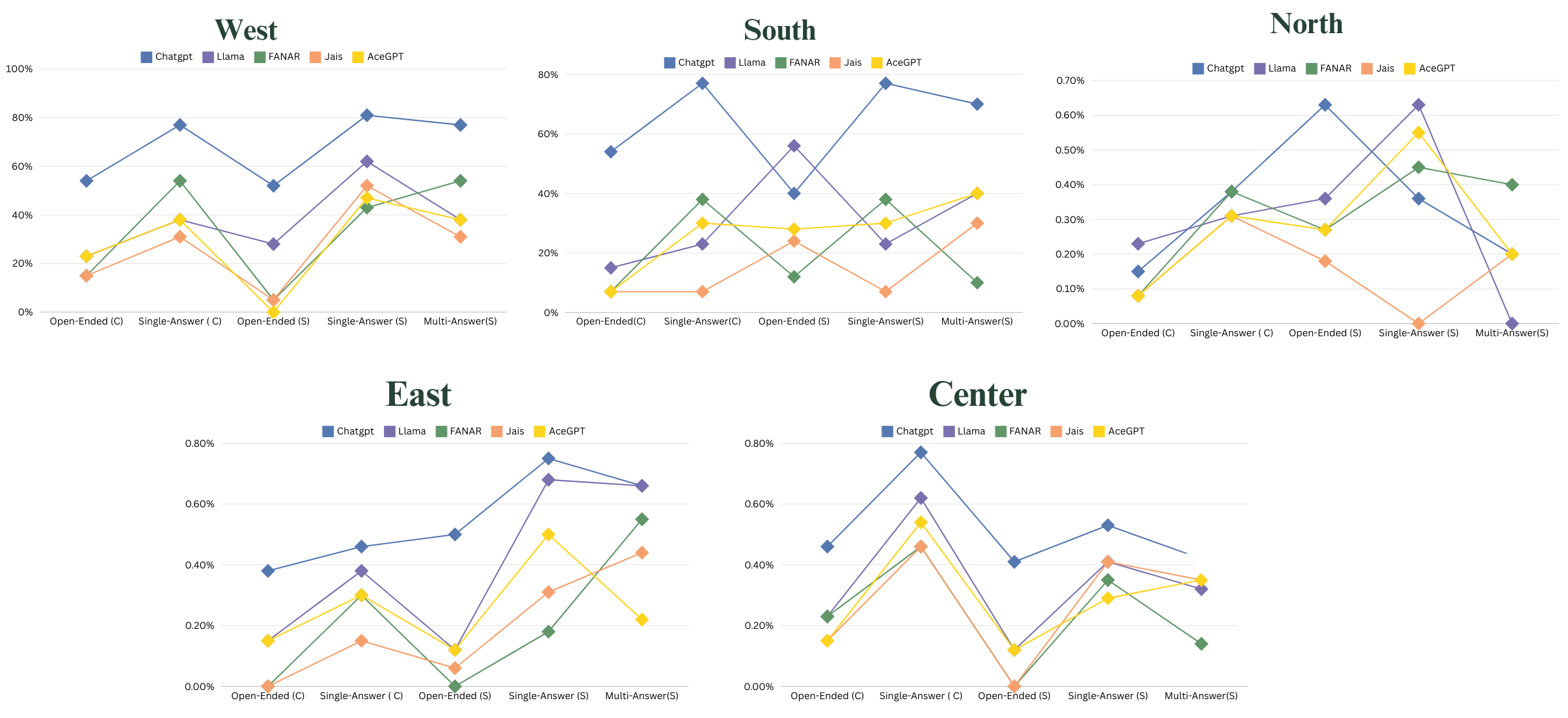}
    \caption{
        Performance comparison of the models across different question formats for multiple regions in Saudi Arabia. The figure includes: \textbf{(1)} Common questions, which involve the most common food, clothing, craft, and other cultural elements, evaluated through both Open-Ended and Single-Answer formats. \textbf{(2)} Specialized cultural questions applicable to a specific region but not to other regions in Saudi Arabia, evaluated through Open-Ended and Single-Answer formats. \textbf{(3)} Specialized cultural questions with more than one correct answer, evaluated through the Multi-Answer format. 
        Each subplot compares the accuracy scores of LLMs across the central, eastern, northern, southern, and western regions.
}
    \label{fig:performance_comparison}
\end{figure}

The analysis, as shown in Figure~\ref{fig:performance_comparison}, reveals that LLMs generally perform better when provided with predefined answer choices, rather than when tasked with generating open-ended responses. In the open-ended format, models are required to produce responses from scratch without any guidance, which often results in lower accuracy due to the inherent complexity of understanding nuanced cultural contexts. On the other hand, when presented with predefined options in the single-answer and multi-answer formats, models benefit from a constrained set of possible answers, making it easier to select the most accurate response. This suggests that predefined answer choices help mitigate the challenges of ambiguity and context interpretation, leading to improved performance across the evaluated models. The increased accuracy in these formats highlights the difficulty LLMs face in generating culturally relevant, contextually appropriate open-ended responses without specific guidance.



The performance of the models in common open-ended questions varies significantly across regions. GPT-4 consistently leads, scoring 54\% in the west and south, while it drops in the center (46\%), east (38\%), and north (15\%). Llama follows, performing best in the North (23\%) but remains weak in the south (15\%) and east (15\%), indicating inconsistencies. FANAR, Jais, and AceGPT all perform poorly, with scores around 7\%-15\% in most regions, except in the center, where FANAR and Llama tie at 23\%.

When handling specialized open-ended questions, the models exhibit a distinct performance trend. GPT-4 continues to lead in most regions, achieving its highest accuracy in the north at 63\% and the west at 52\%. However, its performance declines in the south (40\%) and center (41\%). Notably, Llama surpasses GPT-4 in the south, achieving 56\% accuracy, which highlights its relative strength in handling specialized responses for that region, despite weaker results in other areas. Meanwhile, FANAR, Jais, and AceGPT perform the worst, frequently scoring near 0\% in the center and east, indicating significant challenges in addressing specialized and knowledge-intensive cultural queries.


For common single-answer questions, GPT-4 consistently outperforms the other models across all regions, achieving an accuracy of 77\% in the west, south, and center, while also maintaining relatively strong performance in the east (46\%) and north (38\%). Llama follows with improved results, particularly excelling in the center (62\%), although its performance weakens in the south (23\%) and north (31\%). In contrast, FANAR, Jais, and AceGPT exhibit considerable inconsistencies, with Jais struggling the most, scoring as low as 7\% in the south and 15\% in the east. Overall, the west and central regions appear to be more conducive to AI-generated single-answer responses, whereas the south and north regions show the highest performance variability across all models.


In specialized single-answer questions, GPT-4 remains the strongest, scoring 81\% in the west and 77\% in the south, but its performance drops in the center (53\%) and north (36\%). Llama shows an improvement that performs particularly well in the north (63\%) and east (68\%), surpassing GPT-4 in those regions. FANAR, Jais, and AceGPT remain inconsistent, with Jais even scoring 0\% in the north, indicating a complete failure to handle some specialized factual questions.


For specialized multi-answer questions, GPT-4 maintains its leading position, achieving the highest scores in the west (77\%), south (70\%), and east (66\%). However, its performance drops significantly in the center (42\%) and north (20\%), indicating challenges in providing multiple correct responses for these regions. Llama shows inconsistent performance, excelling in the east (66\%) but scoring 0\% in the north, highlighting a clear regional disparity. FANAR and Jais struggle in most regions, although FANAR performs relatively well in the east (55\%) while dropping to 40\% in the north. AceGPT also exhibits inconsistency, with scores ranging between 20\% and 40\% between regions.

The above discussion demonstrates that the type of question heavily influences the performance of the models. Open-ended questions are the most challenging, especially in specialized contexts requiring deeper reasoning. Single-answer questions are generally easier, with specialized ones demanding domain-specific knowledge. Multi-answer questions are the hardest, requiring both accuracy and the ability to identify multiple valid responses.

\subsection{Comparative Evaluation of Cultural Categories}
The evaluation of cultural knowledge across multiple models offers valuable insights into their ability to comprehend diverse aspects of regional traditions. This analysis assesses the performance of the five models across eight cultural categories: food, clothing, language \& communication, entertainment, celebrations, dating, crafts, and architectural elements. By comparing accuracy scores across these categories, the analysis highlights both areas where the models demonstrate strong understanding and those where they encounter difficulties, thereby identifying critical gaps in their cultural awareness.

\begin{figure}[ht]
    \centering
    \includegraphics[width=\textwidth]{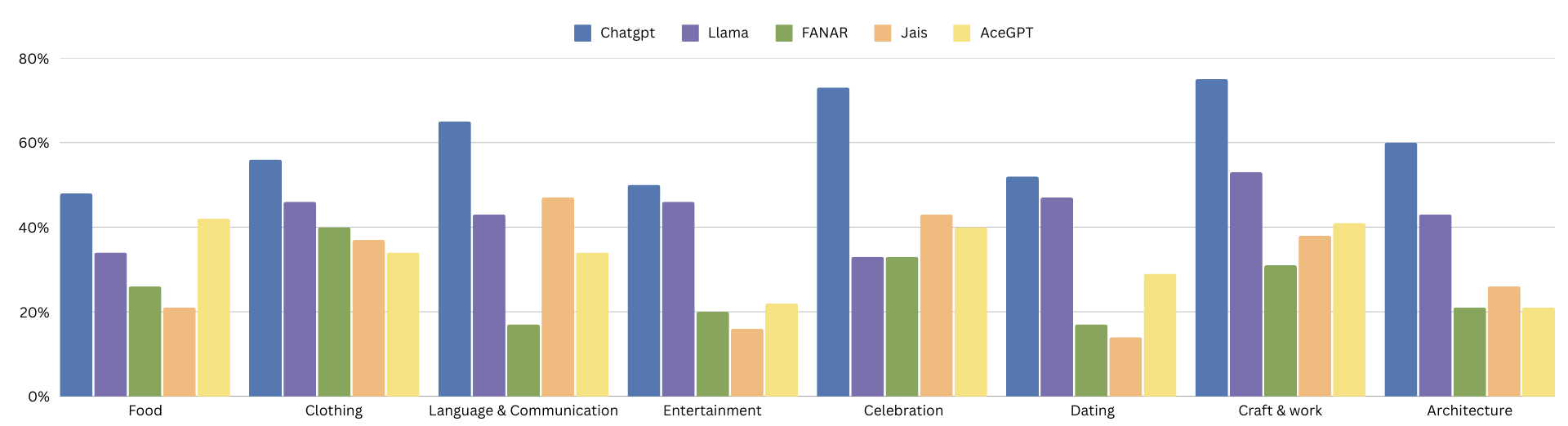}
    \caption{Accuracy comparison of the five  models across eight cultural categories. The figure illustrates variations in model performance for food, clothing, language, entertainment, celebrations, dating, crafts, and architecture cultural aspects, highlighting differences in recognition accuracy across different cultural domains.}
    \label{fig:grouped_bar_chart}
\end{figure}

Across all cultural categories, crafts and work achieved the highest accuracy among the evaluated models, with GPT-4 leading at 75\%, followed by Llama at 53\% as seen in Figure~\ref{fig:grouped_bar_chart}. The celebrations category also yielded relatively strong results, with GPT-4 reaching 73\%, while the other models showed moderate performance. Similarly, language and clothing demonstrated relatively high results, with GPT-4 achieving 65\% and 56\% accuracy, respectively. These findings suggest that models tend to perform better in these more structured cultural categories, maintaining relatively consistent accuracy across different systems.

In contrast, dating customs emerged as the most challenging category, with Jais achieving only 14\% accuracy and FANAR reaching 17\%, while GPT-4—the top-performing model—managed only 52\%. The entertainment category also exhibited weak performance, with Jais scoring 16\% and FANAR 20\%. These results indicate that models face greater difficulty in these less structured or more nuanced cultural domains, with performance varying considerably across models.


\subsection{Case Study: Analysis of LLM Responses on the SaudiCulture Dataset}

Our case study examines the responses generated by five LLMs, evaluating their performance on culturally specific questions from the SaudiCulture dataset. Table ~\ref{tab:my-table1} presents an illustration of these questions alongside their corresponding responses. In the western region example, a question regarding a pre-Ramadan gathering revealed common cultural misinterpretations. GPT-4 and Llama identified this gathering as “Ghabqa,” a term more closely associated with broader Gulf culture rather than the specific cultural context of this Saudi region. Similarly, in response to a question about traditional street food, the correct answer, “Balila,” was frequently overlooked, with several models instead selecting “Shawarma.” While shawarma is a well-known dish in Arabian cuisine, it is not typically classified as a street food in the specified region. This misclassification likely stems from the widespread recognition of shawarma, leading the models to prioritize a more generalized response over a region-specific culinary tradition.

In the southern region, an open-ended question about traditional Tehama clothing resulted in incorrect responses from all four models. Another example from this region involves a multi-answer question about the traditional women's garment, "Al-Mukammam," for which the correct response is "Najran." In this instance, the models were tasked with identifying the region where "Al-Mukammam" is traditionally worn. GPT-4, Llama, and Jais correctly identified "Najran," demonstrating a stronger grasp of regional specificity.

\begin{landscape}
\begin{longtable}{lllcccccc}
\caption{An example of SuadiCulture dataset evaluating different LLMs on questions in various regions of Saudi
Arabia, categorized by  topics and question type. The table compares model responses against the ground truth, with correct answers
highlighted in green and incorrect ones in red. Q: refers to question. GT: refers to ground-truth. L.\&C.: refers to langauge and communication.}
\label{tab:my-table1}\\
\hline
\textbf{Topic} &
  \textbf{\begin{tabular}[c]{@{}l@{}}Q-Type\end{tabular}} &
  \textbf{Q} &
  \textbf{\begin{tabular}[c]{@{}l@{}}GT\end{tabular}} &
  \textbf{GPT-4} &
  \textbf{Llama} &
  \textbf{FANAR} &
  \textbf{Jais} &
  \textbf{AceGPT} \\ \hline
\endhead
\multicolumn{9}{c}{\cellcolor[HTML]{EFEFEF}The Westren Region of Saudi Arabia} \\ \hline
\begin{tabular}[c]{@{}l@{}}Celeb-\\ ration\end{tabular} &
  \begin{tabular}[c]{@{}l@{}}Open-\\ End\end{tabular} &
  \begin{tabular}[c]{@{}l@{}}What's the name \\ of a family gathering\\ that happens before \\ Ramadan?\end{tabular} &
  Shabna &
  \cellcolor[HTML]{F4CCCC}Ghabgah &
  \cellcolor[HTML]{F4CCCC}Ghabqa &
  \cellcolor[HTML]{F4CCCC}- &
  \cellcolor[HTML]{F4CCCC}\begin{tabular}[c]{@{}l@{}}Pre-\\ Ramadan\end{tabular} &
  \cellcolor[HTML]{F4CCCC}Qahwah \\ \hline
Food &
  \begin{tabular}[c]{@{}l@{}}Single-\\ Answer\end{tabular} &
  \begin{tabular}[c]{@{}l@{}}What is the most \\ common street \\ food in the \\ West of \\ Saudi Arabia?\\ Choices:\\ \\ A. Tanoor bread \\ B. balila\\ C. Falfal\\ D. Shawarma\end{tabular} &
  Balila &
  \cellcolor[HTML]{D9EAD3}Balila &
  \cellcolor[HTML]{F4CCCC}\begin{tabular}[c]{@{}l@{}}Shaw\\ arma\end{tabular} &
  \cellcolor[HTML]{F4CCCC}\begin{tabular}[c]{@{}l@{}}Shaw-\\ arma\end{tabular} &
  \cellcolor[HTML]{F4CCCC}\begin{tabular}[c]{@{}l@{}}Shaw-\\ arma\end{tabular} &
  \cellcolor[HTML]{F4CCCC}\begin{tabular}[c]{@{}l@{}}Shaw-\\ arma\end{tabular} \\ \hline
\multicolumn{9}{c}{\cellcolor[HTML]{EFEFEF}The Southern Region of Saudi Arabia} \\ \hline
Clothing &
  \begin{tabular}[c]{@{}l@{}}Open-\\ End\end{tabular} &
  \begin{tabular}[c]{@{}l@{}}What is the name of \\ the traditional clothing\\  in Tehama, located in\\  the southern region of \\ Saudi Arabia?\end{tabular} &
  Wizra &
  \cellcolor[HTML]{F4CCCC}Muhtash &
  \cellcolor[HTML]{F4CCCC}Dishdasha &
  \cellcolor[HTML]{F4CCCC}Thawb &
  \cellcolor[HTML]{F4CCCC}Thawb &
  \cellcolor[HTML]{F4CCCC}- \\ \hline
Clothing &
  \begin{tabular}[c]{@{}l@{}}Multipel-\\ answers\end{tabular} &
  \begin{tabular}[c]{@{}l@{}}In which region of \\ Saudi Arabia is\\ "Al-Mukammam" \\ considered a traditional\\  women’s garment?\\ \\ 1. Najran\\ 2. Mecca\\ 3. Hail\\ 4. Tabuk\\ Choices:\\ \\ (1)\\ (1 and 2)\\ ( 2 and 3)\\ (All)\end{tabular} &
  Najran &
  \multicolumn{1}{r}{\cellcolor[HTML]{D9EAD3}(1)} &
  \cellcolor[HTML]{D9EAD3}(1) &
  \cellcolor[HTML]{F4CCCC}(1and 2) &
  \cellcolor[HTML]{D9EAD3}(1) &
  \cellcolor[HTML]{F4CCCC}(2 and 3) \\ \hline
\multicolumn{9}{c}{\cellcolor[HTML]{EFEFEF}The Eastern Region of Saudi Arabia} \\ \hline
\begin{tabular}[c]{@{}l@{}}Enterta-\\ inment\end{tabular} &
  \begin{tabular}[c]{@{}l@{}}Open-\\ End\end{tabular} &
  \begin{tabular}[c]{@{}l@{}}What is the traditional\\ sea folklore in the\\ Eastern region of\\ Saudi Arabia, \\ especially in Qatif?"\end{tabular} &
  Alsinkaly &
  \multicolumn{1}{r}{\cellcolor[HTML]{F4CCCC}Alnahma} &
  \cellcolor[HTML]{F4CCCC}Al-Bahri &
  \cellcolor[HTML]{F4CCCC}\begin{tabular}[c]{@{}l@{}}Sand \\ Boatman\end{tabular} &
  \cellcolor[HTML]{F4CCCC}\begin{tabular}[c]{@{}l@{}}Nautical \\ Songs,\\ Maritime \\ Ballads\end{tabular} &
  \cellcolor[HTML]{F4CCCC}Dhufar \\ \hline
\begin{tabular}[c]{@{}l@{}}Enterta-\\ inment\end{tabular} &
  \begin{tabular}[c]{@{}l@{}}Multipel-\\ answers\end{tabular} &
  \begin{tabular}[c]{@{}l@{}}Which of the following \\ are styles of the\\  "Laybooni" art?\\ \\ 1. Al-Fajri and \\ Al-Bahri\\ 2. Al-Haddadi and \\ Al-Adasani\\ 3. Al-Makhlof and \\ Al-Hassawi\\ 4. Al-Najdi and\\  Al-Hejazi\\ \\ Choices:\\ \\ (1 and 2)\\ (1, 2, and 3)\\ (2 and 4)\\ (All)\end{tabular} &
  (1, 2 and 3) &
  \cellcolor[HTML]{F4CCCC}(1 and 2) &
  \cellcolor[HTML]{D9EAD3}\begin{tabular}[c]{@{}l@{}}(1, 2 and-\\  3)\end{tabular} &
  \cellcolor[HTML]{D9EAD3}{\color[HTML]{071437} \begin{tabular}[c]{@{}l@{}}(1, 2, -\\ and 3)\end{tabular}} &
  \cellcolor[HTML]{F4CCCC}(1 and 2). &
  \cellcolor[HTML]{F4CCCC}(All) \\ \hline
\multicolumn{9}{c}{\cellcolor[HTML]{EFEFEF}The Northern Region of Saudi Arabia} \\ \hline
L.\&C. &
  \begin{tabular}[c]{@{}l@{}}Open-\\ End\end{tabular} &
  \begin{tabular}[c]{@{}l@{}}Is it acceptable for the\\ host to eat with the \\ guests in the North\\  of Saudi Arabia?\end{tabular} &
  \begin{tabular}[c]{@{}l@{}}Not \\ acceptable\end{tabular} &
  \cellcolor[HTML]{F4CCCC}\begin{tabular}[c]{@{}l@{}}Accept-\\ able\end{tabular} &
  \cellcolor[HTML]{D9EAD3}\begin{tabular}[c]{@{}l@{}}Not \\ acceptable\end{tabular} &
  \cellcolor[HTML]{F4CCCC}Yes &
  \cellcolor[HTML]{F4CCCC}Yes &
  \cellcolor[HTML]{D9EAD3}\begin{tabular}[c]{@{}l@{}}Not \\ accep-\\ table\end{tabular} \\ \hline
\begin{tabular}[c]{@{}l@{}}Enterta-\\ inment\end{tabular} &
  \begin{tabular}[c]{@{}l@{}}Single-\\ Answer\end{tabular} &
  \begin{tabular}[c]{@{}l@{}}What is a traditional \\ game, similar to bowling, \\ commonly known in\\ the northern region \\ of Saudi Arabia?\\  \\ Choices:\\ \\ A. Al-Fashaq\\ B. Al-Bakkoura\\ C. Taq Taqiyyah\\ D. Sabaa Al-Hajar\end{tabular} &
  Al-Fashaq &
  \cellcolor[HTML]{F4CCCC}\begin{tabular}[c]{@{}l@{}}Al-Bakk-\\ oura\end{tabular} &
  \cellcolor[HTML]{F4CCCC}\begin{tabular}[c]{@{}l@{}}Sabaa \\ Al-Hajar\end{tabular} &
  \cellcolor[HTML]{F4CCCC}Bakkoura &
  \cellcolor[HTML]{F4CCCC}\begin{tabular}[c]{@{}l@{}}Taq\\  Taqi-\\yyah\end{tabular} &
  \cellcolor[HTML]{F4CCCC}\begin{tabular}[c]{@{}l@{}}Taq \\ Taqi-\\yyah\end{tabular} \\ \hline
\multicolumn{9}{c}{\cellcolor[HTML]{EFEFEF}The Central Region of Saudi Arabia} \\ \hline
Food &
  \begin{tabular}[c]{@{}l@{}}Open-\\ End\end{tabular} &
  \begin{tabular}[c]{@{}l@{}}What is the traditional\\  winter dish made from \\ dates and brown bread\\  in Al-Qassim?\end{tabular} &
  Al-Hanini &
  \cellcolor[HTML]{F4CCCC}Al-Kleija &
  \cellcolor[HTML]{F4CCCC}Umma &
  \cellcolor[HTML]{F4CCCC}{\color[HTML]{071437} Mishmash} &
  \cellcolor[HTML]{F4CCCC}Harees &
  \cellcolor[HTML]{F4CCCC}Harirah \\ \hline
\begin{tabular}[c]{@{}l@{}}Enterta-\\ inment\end{tabular} &
  \begin{tabular}[c]{@{}l@{}}Single-\\ Answer\end{tabular} &
  \begin{tabular}[c]{@{}l@{}}What is the main \\ musical mode of \\ Nijdi Ardah poems?\\ \\ Choices:\\ \\ A. Hijaz \\ B. Saba \\ C. Sika\\  D. Nahawand\end{tabular} &
  Sika &
  \cellcolor[HTML]{F4CCCC}Hijaz &
  \cellcolor[HTML]{F4CCCC}Hijaz &
  \cellcolor[HTML]{F4CCCC}Hijaz &
  \cellcolor[HTML]{D9EAD3}Sika &
  \cellcolor[HTML]{F4CCCC}Hijaz \\ \hline
\multicolumn{9}{c}{\cellcolor[HTML]{EFEFEF}General Qs} \\ \hline
Dating &
  \begin{tabular}[c]{@{}l@{}}Open-\\ End\end{tabular} &
  \begin{tabular}[c]{@{}l@{}}What is the name\\  of the tradition when\\  the groom sees the \\ bride for the first time?\end{tabular} &
  Alshwfa &
  \cellcolor[HTML]{F4CCCC}Al-Milkah &
  \cellcolor[HTML]{F4CCCC}\begin{tabular}[c]{@{}l@{}}Al-Zawaj \\ Al-Mutlaq\end{tabular} &
  \cellcolor[HTML]{F4CCCC}Idda &
  \cellcolor[HTML]{F4CCCC}Laghrib &
  \cellcolor[HTML]{F4CCCC}Khatb. \\ \hline
L.\&C. &
  \begin{tabular}[c]{@{}l@{}}Open-\\ End\end{tabular} &
  \begin{tabular}[c]{@{}l@{}}What does the \\ gesture of throwing\\  the shemagh at the\\  other person signify?\end{tabular} &
  \begin{tabular}[c]{@{}l@{}}Asking him a\\  favor, like\\  having dinner \\ with him or \\ feel proud \\ of him\end{tabular} &
  \cellcolor[HTML]{F4CCCC}\begin{tabular}[c]{@{}l@{}}playful \\ teasing \\ or a jovial \\ gesture.\end{tabular} &
  \cellcolor[HTML]{F4CCCC}\begin{tabular}[c]{@{}l@{}}Disrespect\\ or insult\end{tabular} &
  \cellcolor[HTML]{F4CCCC}\begin{tabular}[c]{@{}l@{}}not \\ a polite or \\ appropriate \\ gesture.\end{tabular} &
  \cellcolor[HTML]{D9EAD3}\begin{tabular}[c]{@{}l@{}}A gesture\\ of respect\end{tabular} &
  \cellcolor[HTML]{D9EAD3}\begin{tabular}[c]{@{}l@{}}A sign \\of appre-\\ciation.
  \end{tabular} \\ \hline
\end{longtable}
\end{landscape}
 In contrast, FANAR and AceGPT provided compound responses—FANAR selected "Najran and Makkah," while AceGPT chose "Makkah and Tabuk"—suggesting a tendency to conflate multiple regions or rely on broader, less precise cultural associations. Given that Makkah and Tabuk are geographically and culturally distinct from Najran, these inaccuracies highlight a significant challenge in maintaining regional fidelity in model-generated responses.

In the northern region, the analysis of social norms and traditional games further highlights the challenges models face in processing culturally nuanced queries. For instance, in a question about dining etiquette, while dining with guests is generally acceptable in other regions of Saudi Arabia. In this region, it is considered inappropriate for the host to eat alongside their guests. However, some models incorrectly suggest otherwise. Similarly, a question regarding a traditional game analogous to bowling reveals notable discrepancies. Certain models incorrectly select responses such as "Al-Bakkoura" or "Taq Taqiyyah" instead of the ground truth, "Al-Fashaq." Notably, "Taq Taqiyyah" refers to a children's game in which players chase the individual who secretly places a hat behind someone, a concept entirely unrelated to bowling. These errors likely stem from the models’ reliance on generalized or widely known cultural references, which fail to capture the distinct practices specific to the northern region.

In the general category, a question about the cultural gesture of throwing the shemagh at another person elicited varied interpretations from the models. The ground truth describes this gesture as a respectful way of requesting a favor, inviting someone to share a meal, or expressing pride in them—indicating a warm and positive social interaction. However, GPT-4 interpreted it as 'playful teasing or a jovial gesture' which, while suggesting a lighthearted act, overlooks the deeper social significance. Llama’s response, labeling it as 'disrespect or insult' starkly contrasts with the intended meaning, suggesting an overgeneralization of non-verbal cues, possibly influenced by contexts where similar gestures carry negative connotations. Similarly, FANAR described the gesture as 'not polite or appropriate' failing to capture its praiseworthy aspects, likely due to exposure to texts that critique or misrepresent such cultural expressions. Conversely, Jais and AceGPT provided responses more closely to the ground truth, with Jais calling it 'a gesture of respect' and AceGPT referring to it as 'a sign of appreciation.' These variations highlight differences in the models’ training data and their ability to contextualize culturally specific non-verbal expressions.

Overall, the differences in model responses highlight the challenges of capturing subtle cultural nuances. Although some models accurately represent cultural gestures, others rely on generalized or incorrect interpretations, often influenced by biases in training data or a lack of region-specific sources. These discrepancies suggest that LLMs may default to widely known cultural narratives rather than recognizing localized traditions and social practices. This limitation is particularly evident in responses to questions involving non-verbal expressions, etiquette, and regionally specific terminology, where incorrect interpretations can distort the intended meaning. Such findings underscore the need for incorporating richly detailed and context-specific cultural information in training datasets to enhance model performance and ensure more accurate and respectful representation of diverse cultural traditions.

\section{Discussion} \label{discussion}

This study introduced SaudiCulture, a novel benchmark designed to evaluate the cultural competency of LLMs within the diverse context of Saudi Arabia, a nation characterized by its rich linguistic and cultural heterogeneity. Our primary motivation was to address the observed gap in LLMs' ability to capture the nuances of diverse cultures, specifically within a non-Western, culturally complex setting. The creation of SaudiCulture, with its comprehensive dataset covering five distinct geographical regions and various cultural domains, aimed to provide a robust tool for assessing LLMs' understanding of Saudi culture.

The evaluation results revealed significant performance disparities among the evaluated models. Notably, GPT-4 consistently outperformed other models, achieving the highest average accuracy, while Jais lagged behind. This aligns with the broader observation that model performance is highly dependent on the quality and representation of cultural data in their training sets \cite{adilazuarda2024towards}. Furthermore, the observed regional variations, with the West emerging as the best-performing region with GPT-4 and the North proving the most challenging with Jais, underscore the uneven distribution of cultural information within the models' knowledge base. This regional difference reflects what other studies have found about algorithmic bias, where underrepresented populations or regions suffer from lower accuracy and potential misrepresentation \cite{ahmad2024generative, mousi2024aradice, sadallah2025commonsense}. The specific challenges faced by all models in handling highly specialized or region-specific questions, especially those requiring multiple correct answers, highlight the limitations of current LLMs in capturing the complexity and multiplicity of cultural knowledge.

SaudiCulture's methodological design, incorporating questions of varying complexity (open-ended, single-choice, and multiple-choice), enabled a nuanced assessment of LLMs' cultural competence.  The observed performance differences across these formats suggest that constrained queries, such as single-choice and multiple-choice, facilitated higher accuracy by narrowing the response space. This aligns with findings from previous studies \cite{chiu2024culturalbench, myung2025blend}, which demonstrate the impact of query structure on retrieval precision. Conversely, open-ended questions and multiple-choice questions with multiple correct answers exposed the models' limitations in generating or recalling nuanced cultural details. This underscores the need for improved knowledge representation techniques that can handle the ambiguity and complexity inherent in cultural data. 

The findings from this study underscore the importance of culturally grounded benchmarks like SaudiCulture for evaluating and enhancing LLMs. The observed performance gaps and regional variations highlight the necessity for more diverse and representative training datasets that accurately reflect the cultural richness and complexity of regions like Saudi Arabia.  Future work should focus on expanding the benchmark to include additional cultural dimensions, dynamic cultural shifts, and a wider range of dialects and social practices. This aligns with the broader goal of building more culturally aware and globally equitable AI systems, capable of understanding and respecting the full spectrum of human cultures. Future research should also explore methods for fine-tuning models with culturally specific datasets and incorporating knowledge graphs to improve contextual understanding and adaptive response generation.

\subsection{Limitations and Ethical Considerations}

Although SaudiCulture marks a significant advancement in evaluating cultural competence in LLMs, several limitations should be acknowledged. First, although the dataset spans five geographical regions and incorporates a diverse set of cultural categories, the cultural landscape of Saudi Arabia is deeply complex and constantly evolving. As a result, some subtle cultural practices, dialectal variations, and micro-cultural distinctions may not be fully captured. Moreover, the performance results reported in this study are tied to the specific set of models evaluated. Different versions, fine-tuned models, or future architectures may exhibit different levels of cultural understanding, limiting the generalizability of current findings. Finally, the dataset’s focus on Saudi Arabia, while offering deep cultural insights, also constrains its applicability to other cultures. Extending this approach to other underrepresented cultural contexts would enhance its relevance to global cultural competence research.

From an ethical standpoint, careful attention was paid to ensuring authentic and respectful representation of Saudi culture throughout the dataset’s creation. However, it is important to recognize that culture is inherently multifaceted, and no dataset can fully capture the breadth of individual experiences and interpretations within a society. Thus, SaudiCulture should be regarded as a foundational resource, not an authoritative or exhaustive account of Saudi culture. Another critical ethical concern relates to bias amplification. Many contemporary LLMs are trained predominantly on Western-centric corpora, which can lead to cultural biases when these models interact with content from non-Western societies. One of the goals of SaudiCulture is to surface and mitigate such biases by providing a culturally grounded evaluation framework. However, the dataset itself, being partly derived from existing cultural platforms and written resources, may also reflect subtle biases present in those sources. This interplay between dataset bias and model bias must be considered when interpreting the evaluation results.

Furthermore, cultural sensitivity played a central role in dataset curation, especially when addressing topics such as religious practices, tribal customs, and gender roles—areas that can be particularly sensitive or contested. To safeguard cultural accuracy and respect, all questions and answers underwent extensive review by native speakers and cultural experts who provided critical feedback and ensured alignment with culturally appropriate representations. Despite these efforts, users of SaudiCulture are encouraged to remain mindful of the socio-cultural sensitivities involved when applying the dataset. Finally, it is important to emphasize that SaudiCulture is designed strictly for research and educational purposes. Its primary objective is to foster a better understanding of how LLMs engage with culturally rich and diverse content.

\section{Conclusion}\label{conc}

LLMs have demonstrated exceptional natural language processing abilities, yet they often struggle to capture the cultural depth and contextual intricacies of different communities. This paper tackles this issue by focusing on Saudi Arabia, a country characterized by its rich linguistic diversity and cultural complexity. To address this gap, we presented \textbf{SaudiCulture}, a new benchmark designed to evaluate the cultural and geographical understanding of LLMs within the Saudi context. SaudiCulture offers a carefully curated dataset of culturally and geographically grounded questions, representing five distinct regions of Saudi Arabia, providing a valuable tool to assess how well LLMs grasp the nuances of Saudi culture. In addition, by incorporating questions of varying complexity - including open-ended questions, single choice, and multiple choice with multiple correct answers — the benchmark enabled a multifaceted evaluation, testing both generative cultural knowledge and factual recall across different question formats.

Our evaluation results highlight significant performance gaps between models, with GPT-4 demonstrating the strongest overall performance, while Jais struggled to accurately capture Saudi cultural knowledge. Interestingly, regional variations also played a critical role, with the west emerging as the most understandable region for the models, while the north posed the greatest challenges. These findings demonstrate the importance of culturally grounded benchmarks in evaluating and improving LLMs, especially when applied to non-Western culturally diverse contexts. SaudiCulture serves as a first step towards filling this evaluation gap, providing a valuable resource for researchers and developers aiming to enhance the cultural competency of future LLMs.

Moving forward, we envision expanding the benchmark to include additional cultural dimensions, dynamic cultural shifts, and a wider range of dialects and social practices. Ultimately, we hope SaudiCulture contributes to building more culturally aware and globally equitable AI systems, capable of understanding and respecting the full spectrum of human cultures.


\section*{Data Availability}
We provided several illustrative examples from our dataset in this paper, and the complete dataset will be made publicly available for research purposes upon acceptance. A detailed description of the data collection process is presented in Section \ref{data}.

\vspace{0.5cm}

\noindent \textbf{CRediT authorship contribution statement}

\textbf{Lama Ayash}: Methodology, Data Curation, Software, Validation, Formal analysis, Visualization, Writing - original draft. \textbf{Hassan Alhuzali}: Conceptualization, Supervision, Validation, Writing - review \& editing. \textbf{Ashwag Alasmari}: Conceptualization, Supervision, Validation, Writing - review \& editing. \textbf{Sultan Aloufi}: Conceptualization \& Validation.

\section*{Declaration of competing interest}
The authors declare that they have no competing interests.


\section*{Acknowledgment}
The authors extend their appreciation to Umm Al-Qura University, Saudi Arabia for funding this research work through grant number: 25UQU4320430GSSR01. The authors would like also to extend their appreciation to the Deanship of Research and Graduate Studies at King Khalid University for funding this work through small group research under grant number RGP1/337/45. The authors also express their sincere gratitude to Walaa Al-Mukhtar, Saja Alhirbish, Khalid Alessa, and Manal Algethami for their invaluable support and insightful contributions during the evaluation of SaudiCulture. 

\textbf{Declaration of generative AI and AI-assisted technologies in the writing process:} During the preparation of this work, we used GPT-4, an AI chatbot developed by OpenAI, to improve our written work. After using this tool/service, we reviewed and edited the content as needed and take full responsibility for the content of the publication.


\begin{appendices}

\end{appendices}

\bibliography{sn-bibliography}

\end{document}